%% file: main.tex
\documentclass[]{article}
\usepackage[letterpaper]{geometry}
\usepackage{mtsummit2015}
\usepackage{times}
\usepackage{url}
\usepackage{latexsym}
\usepackage{natbib}
\usepackage{layout}

\usepackage{graphicx}

\usepackage{algorithm}
\usepackage[noend]{algpseudocode}

\usepackage{url}
\usepackage{subfigure}
\usepackage{array}
\usepackage{sparklines}
\usepackage{adjustbox}
\usepackage{amsmath,amsfonts,amssymb,amsthm,mathtools}
\usepackage[draft]{hyperref}
\usepackage{enumerate}
\usepackage{bm}
\usepackage[detect-all]{siunitx}
\usepackage{paralist}
\usepackage{color}
\usepackage{scrextend}
\usepackage{multirow} 
\usepackage{tabu}

\input{defns}

\usepackage[english]{babel}
\usepackage[utf8]{inputenc}
\usepackage{amsmath}
\usepackage[colorinlistoftodos]{todonotes}

\title{Context Models for OOV Word Translation in Low-Resource Languages}

\author{\name{\bf Angli Liu} \hfill  \addr{anglil@cs.washington.edu}\\
        \addr{Department of Computer Science, University of Washington,
        Seattle, WA, 98195}
\AND
       \name{\bf Katrin Kirchhoff} \hfill \addr{kk2@u.washington.edu}\\
        \addr{Department of Electrical Engineering, University of Washington,
        Seattle, WA, 98195}
}
\date{\today}

\begin{document}
\maketitle

\begin{abstract} 
Out-of-vocabulary word translation is a major problem for the translation of low-resource languages that suffer from a lack of parallel training data. This paper evaluates the contributions of target-language context models towards the translation of OOV words, specifically in those cases where OOV translations are derived from external knowledge sources, such as dictionaries. We develop both neural and non-neural context models and evaluate them within both phrase-based and self-attention based neural machine translation systems. Our results show that neural language models that integrate additional context beyond the current sentence are the most effective in disambiguating possible OOV word translations. We present an efficient second-pass lattice-rescoring method for wide-context neural language models and demonstrate performance improvements over state-of-the-art self-attention based neural MT systems in five out of six low-resource language pairs.  
\end{abstract}

\input{intro}

\input{prior_work}

\input{method}

\input{data}
\input{experiments}
\input{conclusion}
\newpage
\bibliographystyle{apalike}
\bibliography{mtsummit2015}

\end{document}

%% file: defns.tex
\newcommand{\comment}[1]{}

\newcommand{\bi}{\begin{list}{$\bullet$}{
    \setlength{\leftmargin}{1.5 em}
    \setlength{\itemsep}{0 pt}
    \setlength{\topsep}{3 pt}
    \setlength{\parsep}{3 pt}
    \setlength{\partopsep}{0 pt}
    \setlength{\labelwidth}{1 em}
    \setlength{\labelsep}{0.5 em}
    \setlength{\parskip}{0cm}  }}
\newcommand{\ei}{\end{list}}

\newcommand{\BE}{\begin{enumerate}}
\newcommand{\EE}{\end{enumerate}}

\newcommand{\initab}{                           
\begin{tabbing}
XXX \= XXXX \= \kill
}
\newcommand{\begpub}{
\begin{quotation}
\noindent
}

\newcommand{\finpub}{
\end{quotation}
}

%% file: intro.tex
\section{Introduction}
\label{intro}
Translation of out-of-vocabulary (OOV) words (words occurring in the test data but not in the training data) is of major importance in statistical machine translation (MT). It is a particularly difficult problem in low-resource languages, i.e., languages for which parallel training data is extremely sparse, requiring recourse to techniques that are complementary to standard statistical machine translation approaches. The approaches described in this paper were developed for scenarios where the training data comprises at most 100k sentences pairs. Most previous studies in this area have focused on how to generate translation candidates for OOV words, either by segmentation into subword units, projection from other languages, or by leveraging external knowledge sources like dictionaries. Often, however, these methods generate multiple candidates for each OOV word, and the MT system is insufficiently trained to choose the appropriate translation according to the context. 

In this paper we address this problem by utilizing more sophisticated context models based on target-language information. In particular, we develop wide-context models that incorporate information from context beyond current sentence boundaries to resolve translation ambiguity. We compare these against models incorporating information from the current sentence only, and evaluate neural models vs.~count-based sentence completion and graph reranking models. All are evaluated within both phrase-based and attention-based neural machine translation models for 6 low-resource language pairs. Our paper makes several contributions:
\begin{itemize}
\item We evaluate recently proposed neural machine translation (NMT) architectures (purely attention-based neural MT) on low-resource languages and show that, contrary to previous results obtained with sequence-to-sequence models, neural MT performs similarly to phrase-based machine translation (PBMT) in these scenarios, without modifications to the basic model.
\item We develop and compare several wide-context target-language based models for translation disambiguation and find that document-level neural language models are the most effective at resolving translation ambiguities.  
\item We present an efficient lattice rescoring algorithm for wide-context neural language models.
\item We compare our approach against directly adding external translation resources to the training data and show that our approach provides small but consistent improvements on five out of six language pairs. 
\end{itemize}
The rest of the paper is organized as follows. Section \ref{sec:prior} discusses prior work on OOV translation. Section \ref{sec:contextmodels} describes our general approach and presents various context models for translation disambiguation. Section \ref{sec:data} describes the datasets and baseline systems. Section \ref{sec:experiments} provides experimental results followed by a final conclusion in Section \ref{sec:conclusion}.

%% file: prior_work.tex
\section{Prior Work}
\label{sec:prior}
Several techniques for OOV word translation have been developed in the past. 
The simplest of these involves copying OOV words from the source sentence to the MT output. This is a reasonable procedure if most OOV words can be assumed to be numbers or named entities that do not require transliteration. In traditional PBMT systems, the unknown words can simply be passed through to the output. NMT models typically map all OOVs (as well as rare words) to a single unknown word token. \cite{luong2014addressing} trained an NMT system using external word alignment information, which allowed the system to output positional information about OOVs, which were then translated using a dictionary trained from parallel data. Working within the context of neural sequence-to-sequence models with attention, \cite{bahdanau2014neural} and \cite{jean2014using} pursued the same strategy, with the exception that alignment information was derived from the attention layer in the neural model rather than an external knowledge source. In \cite{gulcehre2016pointing} a pointer model was used that can perform both copying and dictionary lookup. None of these studies address the problem of translation ambiguities resulting from added external knowledge sources. In truly low-resource languages, a dictionary obtained from the parallel training data will not have sufficient information to translate OOVs in the test data, as most of these will never have occurred in the training data. External dictionaries could be used in this case, which however requires a principled method of choosing between different translations.

An alternative strategy to address the rare and unknown word problem is to use subword units, i.e., the original text is segmented into chunks of characters, individual characters, or byte sequences. In \cite{chung2016character}, a pure character-level decoder is used while \cite{luong2016achieving} use a mixed model where the word-level decoder can fall back on the character-level decoder. The byte-pair encoding (BPE) approach of \cite{sennrich2015neural} segments the input text into subword units based on an iterative merging of frequent character n-grams and a fixed upper size of the subword inventory. 
The main motivation given for the subword unit approach is that often a transparent relationship exists between OOVs and other known words: compound words and morphological variants can benefit from substantial overlap with other words in the same language, and cognates and named entities benefit from cross-lingual overlap. However, in resource-poor settings a substantial percentage of OOVs has no overt relationship with other words; instead, genuinely novel translations must be produced for words that were never seen and that are unrelated to other words. 

A third class of approaches tries to leverage cognates and lexical borrowing.  \cite{tsvetkov2015lexicon} show that OOV words in low-resource languages that are loan words from a high-resource language can be translated via the high-resource language. However, the translation of OOV words in that work uses a fixed lexicon, not taking possibly multiple candidates into consideration. 
Finally, other studies have tried to exploit additional monolingual data in the source and/or target language. In \cite{irvine2013} new translation pairs were induced from monolingual corpora based on predictive features such as document timestamps, topic features, word frequency, and orthographic features. \cite{saluja2014graph} and \cite{zhao2015learning} explored the possibility of extracting features from extra monolingual corpora to help cover untranslated phrases. Specifically, \cite{saluja2014graph} induced new translation rules from monolingual data with a semi-supervised algorithm. \cite{zhao2015learning} obtained translation rules for OOV words based on phrases with similar continuous representations for which a translation is known. 

Most of the studies described above have focused on neural MT for language pairs with sufficient training data. Recent work on OOV translation for low-resource languages includes \cite{gujral16}, where a combination of approaches (surface and word-embedding based word similarity, transliteration) is used to generate multiple translation candidates for each OOV to improve phrase-based MT. The choice of a particular translation is then made either by a target-side language  model or by the translation model itself through a secondary phrase table enriched with OOV-specific features.

%% file: method.tex
\section{OOV Disambiguation With Context Models}
\label{sec:contextmodels}
\begin{figure}[t]
\centering
\includegraphics[width=9cm, height=6cm]{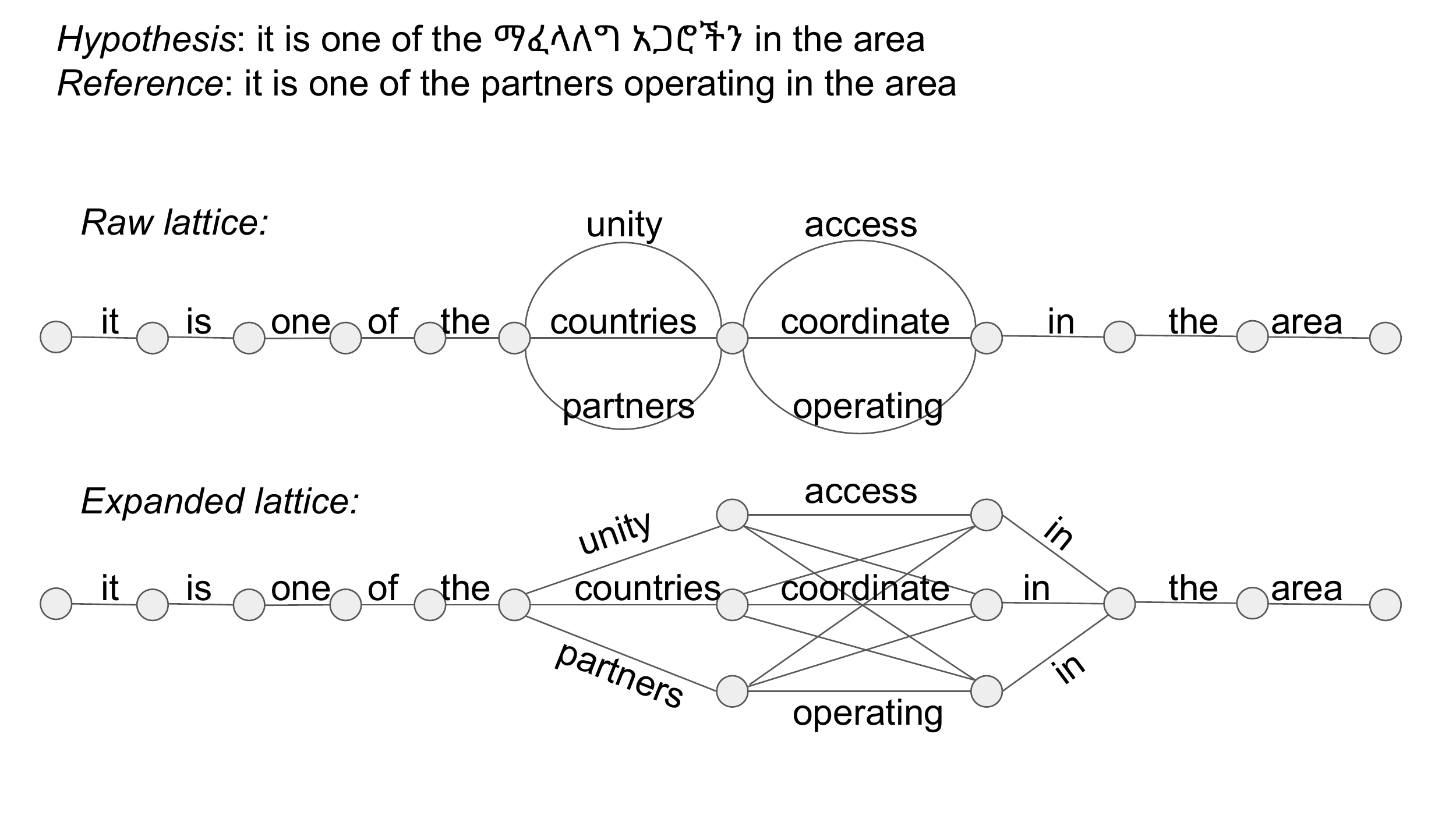}
\caption{Compressed and expanded lattice representations of an MT hypothesis enriched with candidate translations of OOV words. }
\label{fig:lattice}
\end{figure}

Our goal is to facilitate the integration of externally generated translation candidates, such as translation dictionaries, by utilizing a larger amount of target-side context information. We adopt a second-pass lattice rescoring approach that is compatible with both phrase-based and neural MT systems (or their combination) and that can accommodate extra monolingual information without increasing the number of parameters of the MT system itself. OOVs in the MT system's output are expanded to all translation options of that word found in our external knowledge sources. Target-language context models, possibly including context from beyond the current sentence boundaries, are then used to assign a score to each possible path in the extended lattice representing a particular combination of OOV translation hypotheses (see Figure \ref{fig:lattice}).
\subsection{Count-Based Models}
We compare several different models for rescoring, the simplest of which is a {\em sentence completion} model. OOV word translation can be formulated as sentence completion problem, where contextual information informs the filling of  blanks in a sentence. \cite{gubbins2013dependency} proposed to use a language model over a dependency tree for this task, whereas \cite{woods2016exploiting} and \cite{roder2015exploring} measure the degree of association between candidate words and other parts of the sentence using mutual information. In the same spirit our model chooses one out of several possible translation options for each OOV slot in the lattice based its average pointwise mutual information (PMI) with surrounding content words in the sentence (stopwords are ignored). PMI between words $x$ and $y$ is computed as: 
\begin{equation}
PMI(x,y) = log\frac{P(x,y)}{P(x)P(y)}
\label{eq:pmi}
\end{equation}
The algorithm chooses one word at a time, proceeding from left to right. The chosen translation becomes part of the context used for computing PMI for the next set of OOV words. Therefore, the entire space of possible combinations of OOV translations is never fully explored but only searched greedily from left to right. Moreover, this method only focuses on content words and ignores word order in the PMI computation.  

The second model is a graph-based reranking model (\cite{mihalcea05,yang2012unsupervised}), where an undirected graph is built over all OOV translation options and content words in a sentence. Graph edges are weighted with the same PMI values as in the sentence completion approach. PageRank (\cite{page99}) is then used to score each option based on 'votes' from connected words, and for each OOV slot, the options with the highest score is chosen. The PageRank score is computed as 
\begin{equation}
R(v_i) = (1-d) + d\cdot \sum_{j \in IN(v_i)}\frac{R(v_j)}{|OUT(v_j)|} 
\label{eq:pagerank}
\end{equation}
where $v$ is a vertex in the graph, $d$ is a damping factor, and $IN()$ and $OUT()$ are in-degree and out-degree of the vertex, respectively. 
The difference to the sentence completion approach is that the entire space of translation combinations is explored globally rather than greedily. The reason we use the two above methods as baselines is that they can be trained cheaply and readily allow the integration of a larger document context by simply extending the list of content words with words from the previous or following sentences.   
\subsection{Neural Models}
We next turn to neural models. At the sentence-level we utilize a recurrent neural language model, specifically a two-layer long-short term memory (LSTM) model. In order to extend the sentence-level LSTM to include information from previous sentences we follow the approach in \cite{ji2015document}, which proposed several variants of document-context language models (DCLMs). Here, we use an attentional DCLM, which enriches a standard recurrent neural network with a context vector aggregating the hidden vectors in the previous sentence. A standard RNN computes the probability over output classes as 
\begin{equation} 
{\bf y}_{s,n} = softmax({\bf W}_o{\bf h}_{s,n} + {\bf b})
\end{equation}
where $s$ is the current sentence, $n$ is the current time step in the sentence, ${\bf h}$ is a hidden vector, ${\bf b}$ is a bias vector, and ${\bf W}$ is a weight matrix. The hidden vector ${\bf h}_{s,n}$ is computed as 
\begin{equation}
{\bf h}_{s,n} = g({\bf h}_{s,n-1},{\bf x}_{s,n})
\end{equation}
where $g$ is a non-linear function (in this case, representing a two-layer LSTM) and $x$ is an input vector (current word embedding). The attentional DCLM adds a context vector $c_{s,n}$ to both the 
hidden and the output layer as follows:
\begin{eqnarray}
{\bf h}_{s,n} = g({\bf h}_{s,n-1}, [{\bf x}^T_{s,n},{\bf c}^T_{s-1,n}]^T)\\
{\bf y}_{s,n} = softmax({\bf W}_o tanh({\bf W}_h {\bf h}_{s,n}
+ {\bf W}_c  {\bf c}_{s-1,n} + \boldsymbol{b}))
\end{eqnarray}
The context vector $c$ is a position-dependent weighted linear combination of all hidden states $1,...,M$ in the previous sentence. 
\begin{equation}
{\bf c}_{s-1,n}= \sum_{m=1}^M \alpha_{m,n}{\bf h}_{s-1,m}
\end{equation}
The attention weights are computed as
\begin{eqnarray}
a_{n,m} = w_{a}^T tanh({\bf W}_{a 1}{\bf h}_{s,n} + {\bf W}_{a 2}{\bf h}_{s-1,m})\\
\alpha_n = softmax(a_n)
\end{eqnarray}
The attention weights $a_{n,m}$ encode the importance of each word in the previous sentence for the current word. 
DCLMs were shown to obtain reductions in perplexity compared to standard and hierarchical recurrent language models (\cite{ji2015document}); however, they were also observed to have a tendency towards overfitting when training data is sparse (\cite{kirchhoff2016unsupervised}).

\begin{figure}[t]
\centering
\includegraphics[width=8cm]{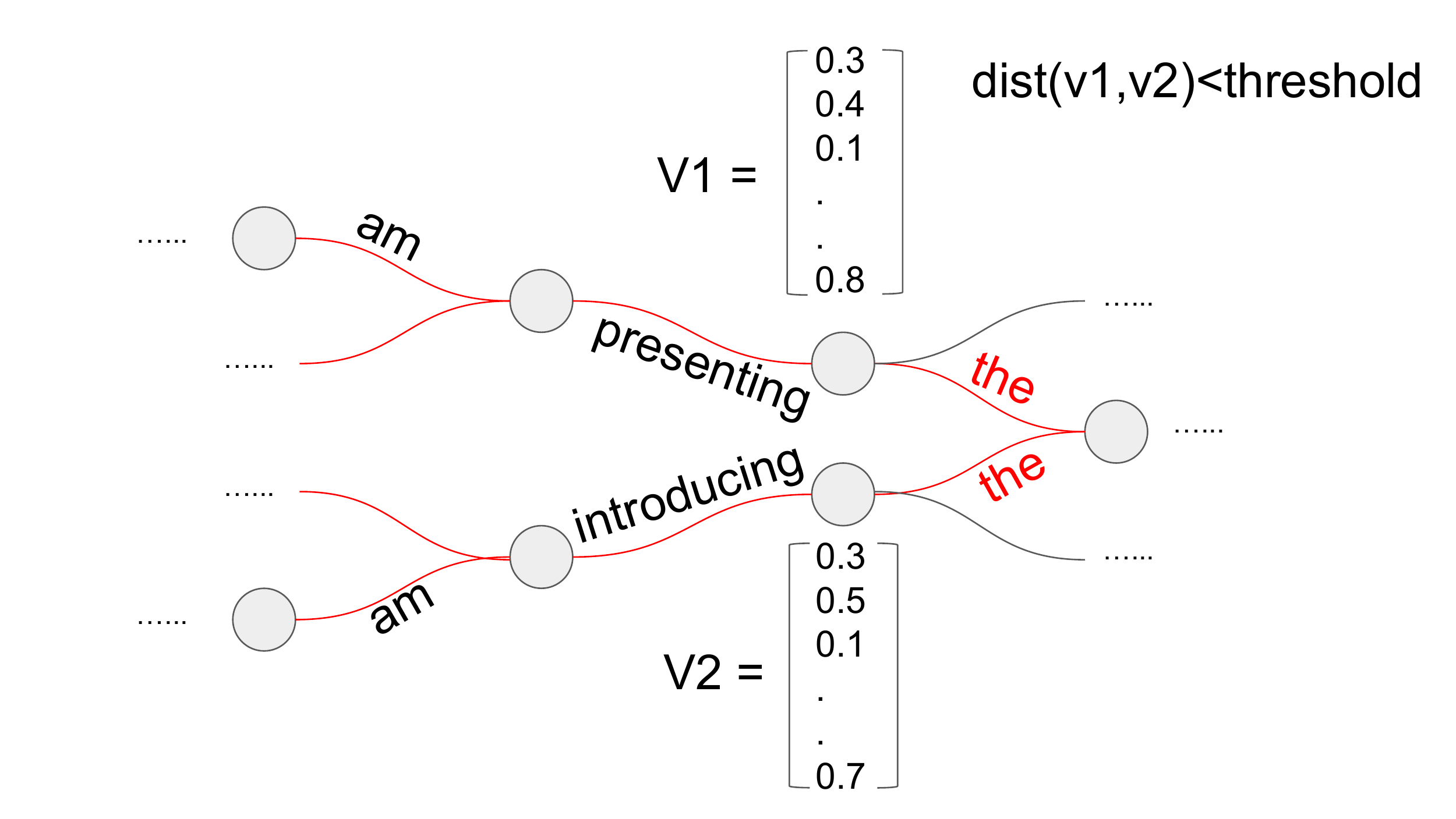}
\caption{\textbf{Lattice path merging.} Paths ending in states whose associated hidden vectors are within a threshold distance of each other are merged.}
\label{fig:merge}
\end{figure}

A different issue in applying neural language models to lattice rescoring is that each path in the lattice defines a different context; however, it is computationally infeasible to exhaustively rescore all paths. The number of OOV words per sentence is typically 3-5 in our tasks, and the number of translation candidates per word may go up to 30. In standard back-off n-gram models, lattice paths are merged based on identical truncated word histories, but this options is not available to us in neural language models where each hidden state encodes the cumulative untruncated history. Inspired by sentence-level lattice rescoring techniques explored in speech recognition (\cite{liu16}), we utilize a document-level lattice rescoring procedure that merges lattice paths based on the similarity of hidden state vectors in the model. The main steps are: 
\begin{enumerate}
\item Start depth-first lattice traversal from the initial node $<s>$. 
\item Use the context matrix $\boldsymbol{c}_{s-1}$ from the previous sentence initialize the hidden representation of the first word $<s>$.
\item At each lattice node, compute the hidden representation and the posterior probability of the word on the incoming arc according to a DCLM. 
\item If there is another lattice path that shares the last word with the current lattice path, and in addition, if the hidden representations of these words fall below some distance threshold $\gamma$, then merge the two paths and update the probability and the hidden representation of the frontier word in the merged lattice path.
\end{enumerate}
A detailed description of the algorithm is provided below in Algorithm \ref{alg:lattice-rescore}.
\begin{algorithm}
\caption{Document-level lattice rescoring}\label{alg:dclm_rescore}
\begin{algorithmic}[1]
\For{each sentence $S$ in document $D$}
\State $L\gets len(S)$
\For{each node $n_i$ in the lattice}
\State initialize its expanded node list $N_i=[]$
\State initialize its expanded outgoing arc list $A_i=[]$
\EndFor
\State $N_0\gets [n'^0_0]$
\State $A_0\gets [a'^0_{01},a'^1_{01},...]$
\For{each expanded node $n'^j_i \in N_i$}
\State create outgoing arcs $a'^0_{i,i+1},a'^1_{i,i+1},...$ according to translation candidates at node $n_{i+1}$
\For{each outgoing arc $a'^k_{i,i+1} \in A_i$}
\State create expanded node $n'^k_{i+1}$
\State $h^k_{i,i+1}\gets$ hidden representation of the DCLM at $a'^k_{i,i+1}$
\State $Pr(a'^k_{i,i+1}|a'^k_{i-1,i},...)\gets$ posterior probability of the lattice path at $a'^k_{i,i+1}$
\If{$\exists a'^l_{i-1,i}\in A_{i-1}$ \textbf{and} $a'^k_{i-1,i}=a'^l_{i-1,i}$ \textbf{and} $d(h^k_{i,i+1},h^l_{i,i+1})<\gamma$}
\If{$Pr(a'^k_{i,i+1}|a'^k_{i-1,i},...)>Pr(a'^l_{i,i+1}|a'^l_{i-1,i},...)$}
\State delete $n'^l_{i+1}$
\State prune the lattice branch that leads to $n'^l_{i+1}$
\Else
\State delete $n'^k_{i+1}$
\State prune the lattice branch that leads to $n'^k_{i+1}$
\EndIf
\EndIf
\EndFor
\EndFor
\State Backtrack from the expanded node $n'^j_L \in N_L$ lattice path that has the highest probability to obtain the decoded sentence.
\EndFor
\end{algorithmic}
\label{alg:lattice-rescore}
\end{algorithm}
As a distance measure we use Euclidean distance between the hidden state vectors. The merging step is illustrated in Figure.~\ref{fig:merge}.
In practice, the procedure can be made computationally efficient by using a cache that maps each explored word to its (possibly multiple) hidden representations and posterior probabilities. In order to find the best path at the end of the traversal, the algorithm looks at all the remaining paths in the cache, finds the one that has the highest log-probability according to the context model, and traces back to the beginning of the path for the entire translation of this sentence. Also, the context matrix for this best path is propagated from the cache to the next sentence in the document for rescoring.

%% file: data.tex
\section{Data and Systems}
\label{sec:data}
\subsection{MT Training Data}
Our experiments are conducted on corpora for six different source languages (Amharic (amh), Uighur (uig), Somali (som), Yoruba (yor),  Vietnamese (vie) and Hausa (hau)), with English as the target language. The corpora were provided as part of the DARPA LORELEI project on low-resource human language technology. Details of the training, development and test set sizes are provided in Table \ref{tab:datastats}. Vocabulary sizes and OOV rates are shown in Table \ref{tab:oovstats}. 
\begin{table}
\begin{center}
	\begin{tabular}{| c | c | c | c | c | c | c |} \hline
     & \multirow{3}{*}{amh} & \multirow{3}{*}{uig} & \multirow{3}{*}{som} & \multirow{3}{*}{yor} & \multirow{3}{*}{hau} & \multirow{3}{*}{vie}\\ 
     & & & & & & \\
    & & & & & & \\ \hline \hline
  	\multirow{3}{*}{train} 
  	& 63,181 / &99,005 / &52,288 / &41,525 / & 43,370 /& 28,686 / \\
    & 5,941 / & 18 / & 8,872 / & 7,538 / & 3,554 / & 439 / \\ 
    & 1,237,172 & 2,587,335 & 1,097,616 & 933,932 & 423,935 & 423,069 \\ \hline
    \multirow{3}{*}{dev} 
    & 992 / &686 / &1,054 / &1,060 / & 957 / & 1,802 / \\
    & 158 / & 3 / & 252 / & 228 / &135 / & 24 / \\
    & 23,085 & 5,156 & 23,113 & 23,080 & 25,782 & 25,730 \\ \hline
    \multirow{3}{*}{test} 
    & 511 / &347 / & 552 / & 594 / &443 / & 196 / \\
    & 90 / & 3 / & 135 /  & 139 /  & 80 / & 7 / \\
    & 11,484 & 2,570 & 11,504 & 11,560 & 4,263 & 4,227 \\ \hline
    \end{tabular}
    \caption{Number of sentence pairs/documents/target language words in the training, development and test sets for each language.}
    \label{tab:datastats}
\end{center}
\end{table}

\subsection{Context Model Training Data}
The training data for the PMI-based context models consists of 4,264,684 Wikipedia articles\footnote{Wikipedia dump of 07/30/2014}. PMI was computed based on the method and implementation described in \cite{roder2015exploring}\footnote{ https://github.com/dice-group/Palmetto}.  
The training data for each DCLM was selected from the Wiki-103 data set (\cite{merity2016pointer}), a collection of 28,475 Wikipedia articles (103M tokens) specifically curated for document-level language modeling. Data was selected separately for each language pair. For each article, a measure of overlap (Jaccard index) was computed between the article's vocabulary and the combined vocabulary extracted from the dev set target references and the one-best MT hypotheses on the test set. All articles with an index higher than $0.1$ were included in the language model training data. The same data was used for training the sentence-level LSTMs, to be able to directly compare the effect of document-level vs.~sentence-level context on OOV disambiguation. 
\subsection{Baseline MT Systems}
Baseline MT systems were developed for these tasks using phrase-based MT and attention-based neural MT (the Transformer model of \cite{vaswani17})\footnote{We used the implementation provided at \url{https://github.com/tensorflow/tensor2tensor}}. The PBMT system was trained using Moses (\cite{koehn2007moses}) and uses a flat phrase-based model with a maximum phrase length of 7, a backoff 4-gram language model trained on the target side of the parallel training data, and a bidirectional reordering model. The log-linear weights were trained using minimum error rate training on the dev set. The Transformer model was trained using a shared byte-pair encoding, resulting in a subword vocabulary of 8,000 word pieces. The hyperparameters of the Transformer models were tuned on the development sets with respect to the number of layers, layer dimensionality, learning rate, and regularization (dropout). The best parameters turned out to be: dropout rate of 0.1 at all layers, a learning rate of 0.2, 2 layers in the encoder and 2 layers in the decoder, and a hidden layer dimensionality of 512. The beam size during decoding is 4. 
Baseline results are shown in Table \ref{tab:baseline-results}. Scoring was done in a case-insensitive fashion against a single reference translation. 

Previous studies of neural sequence-to-sequence models for resource-poor scenarios (e.g., \cite{koehn17}) have found that PBMT models performed significantly better on low-resource languages unless the NMT models were enriched with additional components, such as a lexical memory (\cite{nguyen2017improving}). By contrast, we find that self-attention based neural MT model performs comparably to PBMT, without any modifications to the basic model. A major contributor to the performance of the NMT models is the segmentation induced by byte-pair encoding, which results in system that outperform PBMT systems in 4 out of 6 language pairs. With a word-based vocabulary, NMT underperforms PBMT in most cases. Not surprisingly, languages with rich concatenative morphology (Amharic, Uighur) seem to benefit most from the subword approach.\footnote{PBMT models trained on the segmented text yielded worse scores than either word-based PBMT or Transformer models, even when using a larger maximum phrase length.} \\
\begin{table}
\begin{center}
\begin{tabular}{| l| c|c|c|} \hline
 Language & PBMT & Transformer & Transformer w/ BPE\\ \hline \hline
 amh  & 16.93/49.6 & 13.15/46.0 & {\bf 17.41/51.3} \\ \hline
 uig  & 7.27/37.8 & 11.33/41.9 &  {\bf 17.22/46.7}\\ \hline
 som & 23.22/57.9 & 20.56/54.9 & {\bf 25.36/59.9} \\ \hline
 yor & 18.22/50.8 & 15.68/49.1 & {\bf 19.22/51.4}\\ \hline
 hau & {\bf 21.86/57.8} & 18.61/54.7 & 21.06/56.4 \\ \hline
 vie & {\bf 25.17/55.6} & 22.83/53.1 & 23.00/54.2 \\ \hline
\end{tabular}
\caption{BLEU/unigram precision on test sets for phrase-based MT (PBMT), Transformer model, and Transformer model with byte-pair encoding (BPE).}
\label{tab:baseline-results}
\end{center}
\end{table}

\subsection{Translation Candidate Generation}
We obtain translation candidates for OOV words from (a) a large collection of web-crawled bilingual lexicons (\cite{rolston16}) and (b) translations projected from related languages through Levenshtein distance based retrieval of words similar in their orthographic form. While the former is a reliable source, the latter method may introduce noise. Table \ref{tab:oovstats} shows the number of OOVs, the coverage obtained by our external sources, accuracy (i.e., percentage of OOVs that have a translation matching the reference translation), and the average number of translation candidates per OOV. For all language pairs except for Uighur (which is morphologically highly complex), at least 80\% of all OOV words receive a translation; however, the accuracy is at most 26\% (note, however, that only a single reference translation was available; thus, synonyms are not counted).  
\begin{table}
\label{tab:vocab}
\begin{center}
	\begin{tabular}{| l| c | c | c | c | c | c |} \hline
                      & amh & uig & som & yor & hau & vie \\ \hline
    Vocab & 149,797 & 25,875 & 102,539 & 54,072 & 44,834 & 17,267\\ \hline
    OOV rate & 18.4\%/ & 32.4\%/ & 14.5\%/ & 13.4\%/ & 8.5\%/ & 10.5\%/ \\ 
    (type)/(token) & 8.8\% & 17.2\% & 4.2\% & 3.1\% & 1.7\% & 6.9\% \\ \hline
   Coverage & 99.8\% & 47.6\% & 85.0\% & 83.7\% & 80.1\% & 86.4\% \\ 
    (type)/(token) & 91.2\% & 82.8\% & 95.8\% & 96.9\% & 98.3\% & 90.3\% \\ \hline
    Accuracy & 5.6\% & 10.5\% & 15.9\% & 8.3\% & 8.8\% & 22.7\% \\ \hline
   \# Candidates & 8.0 & 22.0 & 15.4 & 18.4 & 20.6 & 28.6 \\ \hline  
    \end{tabular} 
    \caption{Vocabulary sizes, OOV rates, coverage, accuracy of external translation sources, and average number of translation candidates per OOV word.}
    \label{tab:oovstats}
\end{center}
\end{table}

%% file: experiments.tex
\section{Experimental Results}
\label{sec:experiments}
As an additional baseline we integrate the externally derived translations by simply adding them to the parallel training data, i.e., each translation pair is treated as an additional 'sentence'. Results are shown in Table \ref{tab:added-data-results}.
\begin{table}
\begin{center}
\begin{tabular}{|c|c|c|} \hline
 & PBMT & Transformer \\ \hline
amh 
  & 17.01 / 50.9 & 17.32 / 50.5 \\ \hline
uig 
 & 7.84 / 42.2& {\bf 20.66 / 51.4}\\ \hline
som 
 & 23.91 / 58.6& {\bf 25.45} / 59.6\\ \hline
yor 
 & 18.35/ 50.6& {\bf 19.87 / 51.9}\\ \hline
hau 
 & {\bf 21.94} / 57.8 & 21.55 / 56.7\\ \hline
vie 
 & 25.15 / 55.4 & 22.54 / 52.4\\ \hline
\end{tabular}
\end{center}
\caption{Results (BLEU/unigram precision) of adding external translations to parallel training data. Boldface numbers are improvements over the best baseline system from Table \ref{tab:baseline-results}.}
\label{tab:added-data-results}
\end{table}
In this scenario, each of the new translation pairs is seen only once and without context; the final translation choice is still made by the MT system that has been trained on the parallel data only. Compared to the baseline results in Table \ref{tab:baseline-results}, we observe only mild improvements, except for Uighur, where the improvement is more pronounced. 

We next conduct a diagnostic experiment designed to evaluate the different context models. To this end we enrich the list of translation candidates for each OOV word with the reference translation, in order to determine to what extent the different models are able to recover the correct translation if it is present in the candidate list. For simplicity we run these experiments on the output of the PBMT system, which, unlike the NMT output, contains the location of OOV words. Translation lattices were constructed from the one-best MT hypothesis and OOV translation candidates. The PMI and Pagerank systems were trained as described in Section \ref{sec:contextmodels}. For PageRank, both sentence-level and document-level versions were trained, where the document context was defined to include the previous 4 sentences.  
We compare these against a sentence-level LSTM and the attentional DCLM described in Section \ref{sec:contextmodels}. The sentence-level LSTM is a unidirectional model with two hidden layers of dimensionality 48. 
The DCLMs have a hidden layer size of 48 and also utilize a context size of 4 sentences. Word embedding vectors in both types of language models are initialized randomly. Both LSTMs and DCLMs were trained with DyNet (\cite{dynet}).\footnote{https://github.com/clab/dynet/tree/v1.1}
The vocabulary for the language models consists of the OOV translation candidates and the words from the one-best MT hypotheses.
A comparison between sentence-level and document-level model perplexities on the dev sets for each language pair is shown in Table \ref{tab:ppl}.
\begin{table}
\begin{center}
	\begin{tabular}{| c | c | c | c | c | c | c |} \hline
Model & amh & uig & som & yor & hau & vie\\ \hline
LSTM & 115.0 & 111.7 & 110.4 & 122.5 & 116.4 & 103.4 \\ \hline
DCLM & 101.7 & 103.1 & 100.3 & 98.6 & 97.3 & 95.4 \\ \hline
\end{tabular}
\end{center}
\caption{Perplexities obtained by LSTM vs.~DCLM on dev sets.}
\label{tab:ppl}
\end{table}

The lattice rescoring results from the diagnostic experiment (Table \ref{tab:context-res1}) show that the attentional DCLM generally works best. The remaining experiments therefore use this model only.
\begin{table}[!ht]
\begin{center}
\begin{tabu}{| c || c | c | c | c | c | c |} \hline
& amh & uig & som & yor & hau & vie \\ \hline \hline
PMI & 
16.68 /  &7.82 /  &22.56 /  &17.91 /  &20.89 /  &25.87 /  \\
sent & 49.3& 37.8& 57.4& 49.6& 57.3& 55.8 \\\hline
PageRank & 
 17.00 /  &7.81 /  & 23.11 /& 18.14 /  & 21.55 /&  25.93 / \\
sent & 50.7 & 38.2 & 58.5 & 50.2& 57.9 & 55.8 \\\hline
PageRank & 
 16.97 / & 8.04 / & {\bf23.13} / & 18.17 / & 21.55 / & 25.95  \\
 doc & 50.8 & 38.9 & 58.6 & 50.2 & 57.9 & 55.9 \\ \hline
 \multirow{2}{*}{LSTM}  &
 17.01 / &9.91 / & 22.98 /& 17.87 /& 21.21 /& 25.97 / \\
 & 50.7 & 49.0 & 59.1 & 50.2 & 57.7 & 56.0 \\\hline
\multirow{2}{*}{DCLM}  &
   {\bf 17.39} /& {\bf 9.96} /&23.03 /& {\bf 18.60} /& {\bf 22.21} &{\bf 26.19 /}\\
 & {\bf 52.9} & {\bf 50.4} & {\bf 59.4} & {\bf 50.8} & {\bf 57.9} & {\bf 56.9} \\ \hline
\end{tabu}
\caption{BLEU/unigram precision for lattice rescoring of PBMT output with reference translation included (diagnostic experiment).}
\label{tab:context-res1}
\end{center}
\end{table}

We next apply the DCLM based rescoring method to our best baseline system, i.e., the Transformer system with BPE. Since this system decomposes all words into word pieces, it is not obvious which part of the output corresponds to an original OOV. We therefore align the PBMT and Transformer outputs, retain only the Transformer output and the aligned OOV slots, and replace OOV slots with their external translation options. We used fastAlign (\cite{dyer2013simple}) for this procedure, treating the Transformer output as source and the PBMT output as target for amh, uig, som, and yor. For vie and hau, we use the PBMT system's output as source since it outperforms the Transformer model. The results are shown in Row 3 in Table \ref{tab:context-res2}. 

With the exception of Uighur, we find that our method slightly but consistently outperforms systems that utilize the external translations as additional training data (Row 2), indicating that contextual information is useful. For further calibration of the results we also provide topline results from an oracle experiment where the correct reference translation was substituted for every OOV slot (Row 4) -- these numbers indicate the maximum possible improvement in BLEU/unigram precision that can be obtained from OOV translation on these tasks. For completeness we also provide the original baseline system scores (Row 1) and results obtained from full system combination (e.g., both the PBMT and the NMT's outputs are represented in the rescoring lattice, in addition to OOV translation options). Not surprisingly, system combination adds further improvements  (except for Uighur), in some cases bringing the overall performance close to the topline. 
\begin{table}[!ht]
\begin{center}
\begin{tabu}{|c| c || c | c | c | c | c | c |} \hline
& Method & amh & uig & som & yor & hau & vie\\ \hline \hline
1& No OOV & 17.41/ & 17.22/ & 25.36/ & 19.22/ & 21.86/ & 25.17/ \\
& rescoring & 51.3 & 46.7 &  59.9 & 51.4& 57.8 & 55.6\\ \hline
2& Add'l & 17.32/ & 20.66/ & 25.45/ &19.87/  & 21.94/ & 25.15/ \\ 
& train data & 50.5 & 51.4 & 59.6 & 51.9 & 57.8 & 55.4 \\ \hline
3& OOV & 17.76 / & 17.33 / & 25.50 / & 19.97 / & 22.42 / & 27.25 / \\ 
& rescoring & 53.8 & 47.1 & 60.2 & 52.8 & 59.9 & 57.7 \\\hline
4& OOV & 18.62 /& 21.48 /& 27.57 /& 21.40 /& 22.77 /& 28.61 /\\ 
& topline & 58.4 & 60.5 & 64.4 & 56.8 & 61.5 & 59.4 \\ \hline
5 & sys. & 18.24 /& 18.10 / & \textbf{27.00 /} & \textbf{20.82 /} & \textbf{22.65 /} & \textbf{28.19} /\\
 & comb. & \textbf{56.1}& 50.9 & \textbf{63.0}& \textbf{55.5}& \textbf{60.7}& \textbf{58.5}\\\hline
\end{tabu}
\caption{BLEU/unigram precision of  (1) baseline system without OOV handling; (2) systems trained with external translations as additional training data; (3) lattice rescoring with context models; (4) oracle; (5) full system combination of PBMT and Transformer outputs plus OOV translation.}
\label{tab:context-res2}
\end{center}
\end{table}
An example of the different system outputs is shown below:\\
{\bf Source sentence (with OOVs in italics):} \\
saraakiisha ayaa sheegaya in qaraxu ka dhacay meel u dhow {\em albaadka} aqalka baarlamaanka , kaddib markii ilaaladu ay rasaas ku furtay {\em baabuurkaasi}\\
{\bf No oov rescoring:}\\
officials said that the explosion took place near the parliament albaadka, after they opened fire on baabuurkaasi\\
{\bf Transformer output:}\\
officials say that the explosion occurred near the house of the parliament after the guards opened fire on that vehicle\\
{\bf After rescoring with context model (our method):}\\
officials said that the explosion took place near the parliament entrance, after they opened fire on kondoo\\
{\bf System combination:}\\
officials said that the explosion occurred near the entrance of the parliament after the guards opened fire on that vehicle\\
{\bf Reference:}\\
officials said the explosion took place near the entrance of the parliament building when guards opened fire on the vehicle\\[0.2cm]
The best baseline system (Transformer with BPE) was able to correctly handle {\em baabuurkaasi} (vehicle) but not {\em albaddka} (entrance), which the rescoring procedure corrected. While this procedure also introduces an incorrect word ({\em kondoo}), rescoring of the lattice representing both the PBMT and Transformer output (system combination) in addition to OOV translations results in the correct output. 

%% file: conclusion.tex
\section{Conclusion}
\label{sec:conclusion}
We have presented an approach towards the resolution of ambiguous translations of OOV words that arise when adding word translation pairs from external knowledge sources to an MT system. Of the different context models proposed, document-context language models with a context including previous sentences were shown to be most effective at identifying the correct translations. Our method showed substantial gains over baseline systems without special OOV handling and small but consistent gains over adding external translations directly to the training data, in five out of six language pairs.
Future work will be concerned with integrating external resources and contextual information directly into neural MT architectures.\\[0.5cm]
\noindent
{\bf Acknowledgements}\\
This paper was funded by DARPA under cooperative agreement no.~HR0011-15-2-0043. The content of the information does not necessarily reflect the position or the policy of the Government, and no official endorsement should be inferred.

%% file: main.bbl
\begin{thebibliography}{}

\bibitem[Bahdanau et~al., 2014]{bahdanau2014neural}
Bahdanau, D., Cho, K., and Bengio, Y. (2014).
\newblock Neural machine translation by jointly learning to align and
  translate.
\newblock {\em arXiv preprint arXiv:1409.0473}.

\bibitem[Chung et~al., 2016]{chung2016character}
Chung, J., Cho, K., and Bengio, Y. (2016).
\newblock A character-level decoder without explicit segmentation for neural
  machine translation.
\newblock {\em arXiv preprint arXiv:1603.06147}.

\bibitem[Dyer et~al., 2013]{dyer2013simple}
Dyer, C., Chahuneau, V., and Smith, N.~A. (2013).
\newblock A simple, fast, and effective reparameterization of ibm model 2.
\newblock Association for Computational Linguistics.

\bibitem[Gubbins and Vlachos, 2013]{gubbins2013dependency}
Gubbins, J. and Vlachos, A. (2013).
\newblock Dependency language models for sentence completion.
\newblock In {\em EMNLP}, volume~13, pages 1405--1410. Citeseer.

\bibitem[Gujral et~al., 2016]{gujral16}
Gujral, B., Khayralla, H., and Koehn, P. (2016).
\newblock Translation of unknown words in low-resource languages.
\newblock In {\em Proceedings of AMTA}.

\bibitem[Gulcehre et~al., 2016]{gulcehre2016pointing}
Gulcehre, C., Ahn, S., Nallapati, R., Zhou, B., and Bengio, Y. (2016).
\newblock Pointing the unknown words.
\newblock {\em arXiv preprint arXiv:1603.08148}.

\bibitem[Irvine and Callison-Burch, 2013]{irvine2013}
Irvine, A. and Callison-Burch, C. (2013).
\newblock Supervised bilingual lexicon induction with multiple monolingual
  signals.
\newblock In {\em Proceedings of HLT-NAACL}, page 518–523.

\bibitem[Jean et~al., 2014]{jean2014using}
Jean, S., Cho, K., Memisevic, R., and Bengio, Y. (2014).
\newblock On using very large target vocabulary for neural machine translation.
\newblock {\em arXiv preprint arXiv:1412.2007}.

\bibitem[Ji et~al., 2015]{ji2015document}
Ji, Y., Cohn, T., Kong, L., Dyer, C., and Eisenstein, J. (2015).
\newblock Document context language models.
\newblock {\em arXiv preprint arXiv:1511.03962}.

\bibitem[Kirchhoff and Turner, 2016]{kirchhoff2016unsupervised}
Kirchhoff, K. and Turner, A.~M. (2016).
\newblock Unsupervised resolution of acronyms and abbreviations in nursing
  notes using document-level context models.
\newblock {\em EMNLP 2016}, page~52.

\bibitem[Koehn et~al., 2007]{koehn2007moses}
Koehn, P., Hoang, H., Birch, A., Callison-Burch, C., Federico, M., Bertoldi,
  N., Cowan, B., Shen, W., Moran, C., Zens, R., et~al. (2007).
\newblock Moses: Open source toolkit for statistical machine translation.
\newblock In {\em Proceedings of the 45th annual meeting of the ACL on
  interactive poster and demonstration sessions}, pages 177--180. Association
  for Computational Linguistics.

\bibitem[Koehn and Knowles, 2017]{koehn17}
Koehn, P. and Knowles, R. (2017).
\newblock Six challenges for neural machine translation.
\newblock In {\em Proceedings of the First Workshop on Neural Machine
  Translation}, pages 28--39, Vancouver. Association for Computational
  Linguistics.

\bibitem[Liu et~al., 2016]{liu16}
Liu, X., Chen, X., Wang, Y., Gales, M., and Woodland, P. (2016).
\newblock Two efficient lattice rescoring methods using recurrent neural
  network language models.
\newblock volume 24(8).

\bibitem[Luong and Manning, 2016]{luong2016achieving}
Luong, M.-T. and Manning, C.~D. (2016).
\newblock Achieving open vocabulary neural machine translation with hybrid
  word-character models.
\newblock {\em arXiv preprint arXiv:1604.00788}.

\bibitem[Luong et~al., 2014]{luong2014addressing}
Luong, M.-T., Sutskever, I., Le, Q.~V., Vinyals, O., and Zaremba, W. (2014).
\newblock Addressing the rare word problem in neural machine translation.
\newblock {\em arXiv preprint arXiv:1410.8206}.

\bibitem[Merity et~al., 2016]{merity2016pointer}
Merity, S., Xiong, C., Bradbury, J., and Socher, R. (2016).
\newblock Pointer sentinel mixture models.
\newblock {\em arXiv preprint arXiv:1609.07843}.

\bibitem[Mihalcea, 2005]{mihalcea05}
Mihalcea, R. (2005).
\newblock Unsupervised large-vocabulary word sense disambiguation with
  graph-based algorithms for sequence data labeling.
\newblock In {\em Proceedings of EMNLP}.

\bibitem[Neubig et~al., 2017]{dynet}
Neubig, G., Dyer, C., Goldberg, Y., Matthews, A., Ammar, W., Anastasopoulos,
  A., Ballesteros, M., Chiang, D., Clothiaux, D., Cohn, T., Duh, K., Faruqui,
  M., Gan, C., Garrette, D., Ji, Y., Kong, L., Kuncoro, A., Kumar, G.,
  Malaviya, C., Michel, P., Oda, Y., Richardson, M., Saphra, N., Swayamdipta,
  S., and Yin, P. (2017).
\newblock Dynet: The dynamic neural network toolkit.
\newblock {\em arXiv preprint arXiv:1701.03980}.

\bibitem[Nguyen and Chiang, 2017]{nguyen2017improving}
Nguyen, T.~Q. and Chiang, D. (2017).
\newblock Improving lexical choice in neural machine translation.
\newblock {\em arXiv preprint arXiv:1710.01329}.

\bibitem[Page et~al., 1999]{page99}
Page, L., Brin, S., Motwani, R., and Winograd, T. (1999).
\newblock The pagerank citation ranking: Bringing order to the web.
\newblock Technical Report 1999-66, Stanford InfoLab.

\bibitem[R{\"o}der et~al., 2015]{roder2015exploring}
R{\"o}der, M., Both, A., and Hinneburg, A. (2015).
\newblock Exploring the space of topic coherence measures.
\newblock In {\em Proceedings of the eighth ACM international conference on Web
  search and data mining}, pages 399--408. ACM.

\bibitem[Rolston and Kirchhoff, 2016]{rolston16}
Rolston, L. and Kirchhoff, K. (2016).
\newblock Collection of bilingual data for lexicon transfer learning.
\newblock Technical Report UW-EE-2016-0001.

\bibitem[Saluja et~al., 2014]{saluja2014graph}
Saluja, A., Hassan, H., Toutanova, K., and Quirk, C. (2014).
\newblock Graph-based semi-supervised learning of translation models from
  monolingual data.
\newblock In {\em ACL (1)}, pages 676--686.

\bibitem[Sennrich et~al., 2015]{sennrich2015neural}
Sennrich, R., Haddow, B., and Birch, A. (2015).
\newblock Neural machine translation of rare words with subword units.
\newblock {\em arXiv preprint arXiv:1508.07909}.

\bibitem[Tsvetkov and Dyer, 2015]{tsvetkov2015lexicon}
Tsvetkov, Y. and Dyer, C. (2015).
\newblock Lexicon stratification for translating out-of-vocabulary words.
\newblock In {\em ACL (2)}, pages 125--131.

\bibitem[Vaswani et~al., 2017]{vaswani17}
Vaswani, A., Shazeer, N., Parmar, N., Uszkoreit, J., Jones, L., Gomez, A.,
  Kaiser, L., and Polosukhin, I. (2017).
\newblock Attention is all you need.
\newblock arXiv:1706.03762.

\bibitem[Woods, 2016]{woods2016exploiting}
Woods, A.~M. (2016).
\newblock Exploiting linguistic features for sentence completion.
\newblock In {\em The 54th Annual Meeting of the Association for Computational
  Linguistics}, page 438.

\bibitem[Yang and Kirchhoff, 2012]{yang2012unsupervised}
Yang, M. and Kirchhoff, K. (2012).
\newblock Unsupervised translation disambiguation for cross-domain statistical
  machine translation.
\newblock In {\em Proceedings of association for machine translation in the
  Americas}.

\bibitem[Zhao et~al., 2015]{zhao2015learning}
Zhao, K., Hassan, H., and Auli, M. (2015).
\newblock Learning translation models from monolingual continuous
  representations.
\newblock In {\em HLT-NAACL}, pages 1527--1536.

\end{thebibliography}
